\documentclass[lettersize,journal]{IEEEtran}
\usepackage{amsmath,amsfonts}
\usepackage{algpseudocode}
\usepackage{algorithmicx}
\usepackage{algorithm}
\usepackage{setspace}
\usepackage{array}
\usepackage[caption=false,font=normalsize,labelfont=sf,textfont=sf]{subfig}
\usepackage{textcomp}
\usepackage{stfloats}
\usepackage{url}
\usepackage{verbatim}
\usepackage{graphicx}
\usepackage{cite}
\usepackage{booktabs}
\usepackage{multirow}

\begin{document}

\title{Unsupervised Manga Character Re-identification via Face-body and Spatial-temporal Associated Clustering}

\author{Zhimin Zhang, Zheng Wang,~\IEEEmembership{Member,~IEEE}, Wei Hu,~\IEEEmembership{Senior~Member,~IEEE}
\IEEEcompsocitemizethanks{
\IEEEcompsocthanksitem \textit{Corresponding authors: Zheng Wang and Wei Hu.}
\IEEEcompsocthanksitem Zhimin Zhang and Zheng Wang are with the School of Computer Science, Wuhan University, No.299, Bayi Road, Wuchang District, Wuhan City, Hubei Province, China (e-mails: \{zhangzhimin,wangzwhu\}@whu.edu.cn).
\IEEEcompsocthanksitem Wei Hu is with Wangxuan Institute of Computer Technology, Peking University, No. 128, Zhongguancun North Street, Beijing, China (e-mail: forhuwei@pku.edu.cn).
}}



\maketitle

\begin{abstract}
In the past few years, there has been a dramatic growth in e-manga (electronic Japanese-style comics). Faced with the booming demand for manga research and the large amount of unlabeled manga data, we raised a new task, called \textit{unsupervised manga character re-identification}. However, the artistic expression and stylistic limitations of manga pose many challenges to the re-identification problem. Inspired by the idea that some content-related features may help clustering, we propose a Face-body and Spatial-temporal Associated Clustering method (FSAC). 
In the face-body combination module, a face-body graph is constructed to solve problems such as exaggeration and deformation in artistic creation by using the integrity of the image. 
In the spatial-temporal relationship correction module, we analyze the appearance features of characters and design a temporal-spatial-related triplet loss to fine-tune the clustering.
Extensive experiments on a manga book dataset with 109 volumes validate the superiority of our method in unsupervised manga character re-identification.
\end{abstract}


\section{Introduction}
\IEEEPARstart{R}{ecent}
years have witnessed increasing attention in cartoon and manga (Japanese-style comics) \cite{mangapopularity}, driven by the strong demands of industrial applications. E-Manga is also becoming more popular as people's reading patterns have changed dramatically. For example, Amazon's Kindle store has over 60,000 e-manga on sale\footnote{Amazon.com Kindle Comic. Retrieved on August 12, 2021, from http://amzn.to/1KD5ZBK}.
Manga character recognition is a crucial task in manga-related research, which plays an important role in areas such as character retrieval and character clustering \cite{sketch,mangaclustering}.

Nevertheless, most existing character recognition methods require large amounts of labeled data \cite{cartoon}, which limits the wide usability and scalability in real-world application scenarios, as it is both expensive and difficult to manually label large datasets.
Unsupervised approaches have broader applicability in the face of a plethora of manga characters, but remain relatively under-explored.

\begin{figure}[t]
\centering\includegraphics[scale=0.25]{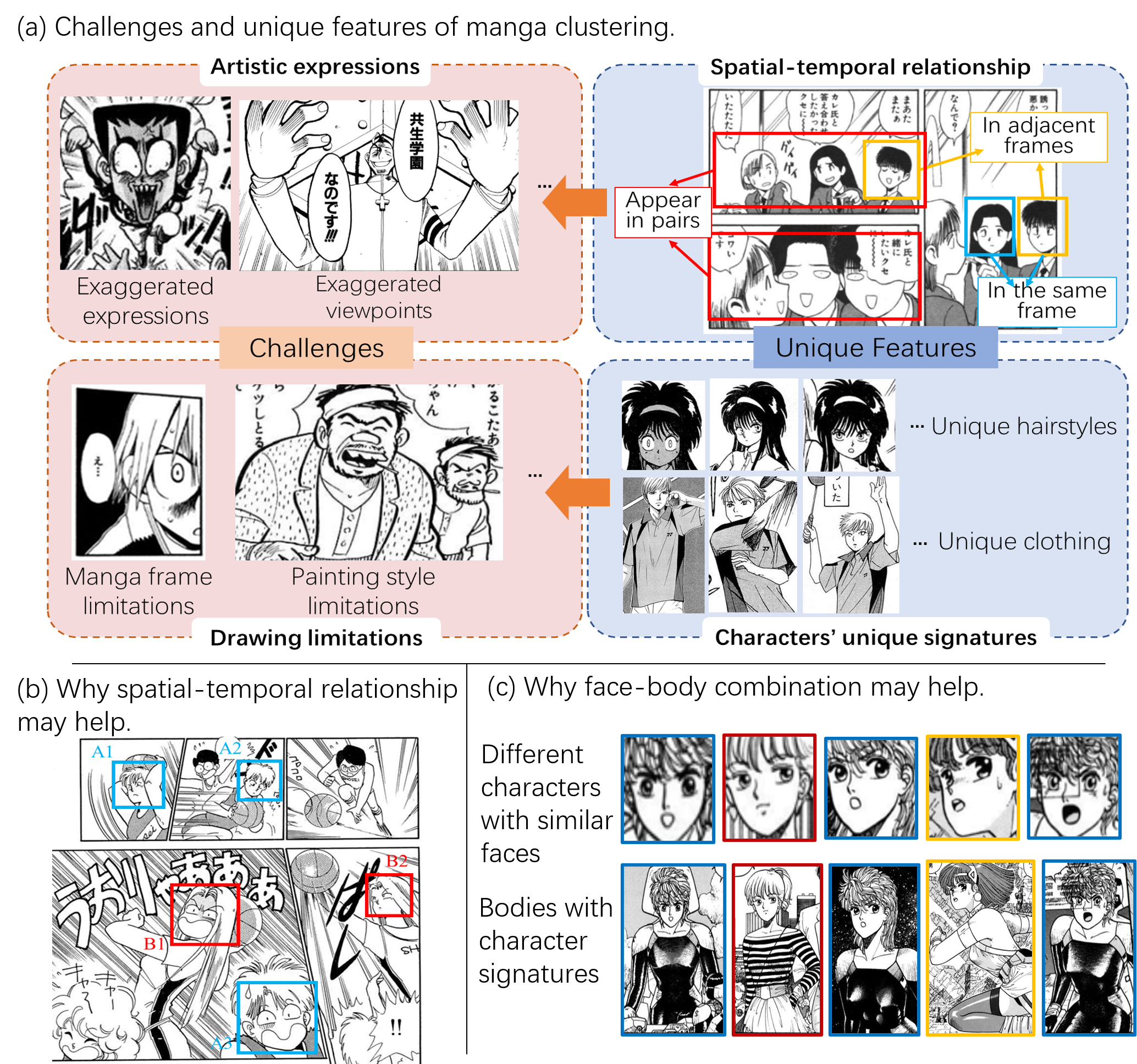}
\caption{(a) Taking advantage of the unique features of manga to make up for the challenges of the transfer task. (b) Why spatial-temporal relationship may help: exaggerated expressions such as B1 and A3 are difficult to recognize, but from the perspective of manga spatial-temporal relationship, A1-A3 and B1-B2 appear in sequences. (c) Why face-body combination may help: due to the limitation of painting style, different characters have similar faces, but different clothes constitute the signatures of the characters.}\label{fig:intro}
\end{figure}

Faced with the booming demand for manga research and the large amount of unlabeled manga data, we raised a new task, called \textit{Unsupervised Manga character Re-identification (or UManga-ReID)}. 
A common strategy for UManga-ReID is the unsupervised domain adaptation approach, which transfers the learned knowledge from the source domain by optimizing with pseudo labels created by clustering algorithms to the target domain \cite{MMT}. 
In view of the intrinsic similarity between the real-person identities and cartoon characters, most related studies \cite{cartoon} adopt real-person data as the source domain and cartoon data as the target domain.
However, although this approach is effective to some extent, it is often plagued by the noise generated by pseudo-labeling. How to effectively reduce the noise and obtain better clustering performance is the key to this problem.

Although there exhibit many similarities between manga and the real world, manga has a more important task, which is narrative.
Some manga-related research has focused on the content of manga.
\cite{amazing} points out that there are certain patterns of transition between manga frames. \cite{cartoon} discusses the influence of context on the accuracy of manga face recognition.
These provide theoretical guidance for improving the clustering performance from the perspective of manga content.

Manga characters are often artistic creations of real people, so transferring models trained on real people data to manga characters is a feasible approach. However, as an art form, manga is expressed in a very different way from the real world, which poses many challenges to the transfer task (see Figure~\ref{fig:intro}(a)):

\begin{itemize}
    \item \textbf{Artistic expressions:} For example, the exaggerated viewpoints and the exaggerated expressions of characters, \textit{etc}.
    \item \textbf{Drawing limitations:} For example, the limitations of the frame cause some characters' bodies to be left out, and the limitations of the drawing style make characters in the same manga book look similar.
\end{itemize}

On the other hand, a more important task of manga is narration, which brings many characteristics that real data do not have and may further help us in the UManga-ReID study:

\begin{itemize}
    \item \textbf{Temporal-spatial relationship:} For example, the same characters tend to appear on adjacent pages one after another, characters in the same frame tend to belong to different identities, some characters tend to appear in pairs, and so on.
     The frame order of the manga tells the contents of the book in chronological order, and we define the relationships on the frame order as \textit{temporal relationships}.
     Moreover, each frame in the manga often represents a scene, and we define the relationships between characters in the same frame as \textit{spatial relationships}.
    \item \textbf{Characters' unique signatures:} Many characters have unique signatures, such as special accessories, hairstyles, costumes, \textit{etc}. These signatures usually do not change in the same manga book.
\end{itemize}

\begin{figure}[t]
\centering\includegraphics[scale=0.58]{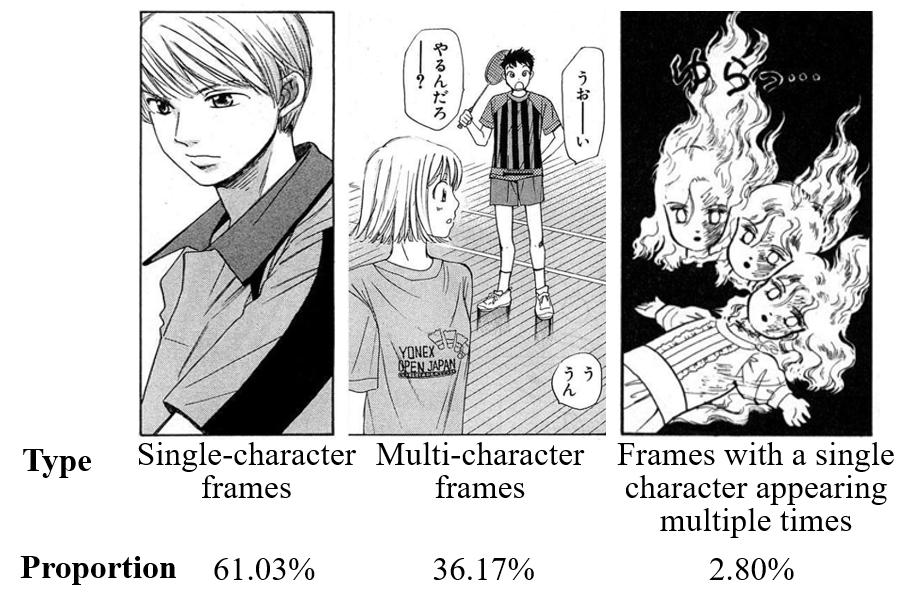}
\caption{
The spatial relationships of frames can be divided into three types: single-character frames, multi-character frames, and frames with a single character appearing multiple times. Their probabilities are shown below. We can see that most of the characters appearing in the same manga frame belong to different identities.}\label{fig:proportion}
\end{figure}

In order to verify the feasibility of our conjecture, we conducted statistics on the spatial-temporal rules of manga characters.
Figure~\ref{fig:proportion} shows the spatial relationship of manga, revealing that characters appearing in the same frame are more likely to belong to different identities.
Table~\ref{tab:proportion} shows the temporal relationship, and the statistical results show that a character has a high probability of appearing in neighboring frames.

We are interested in knowing whether the unique features of manga can be exploited to compensate for the challenges of transfer tasks and thus achieve better results in UManga-ReID. Based on the above analysis, we have further conjectures: On the one hand, there are some potential patterns among the appearance order of manga book characters that enable us to constrain and complement the clustering from the perspective of spatial-temporal relationships when facing exaggerated or omitted artistic representations. 
On the other hand, different characters in the same manga book often look similar due to the constraints of the manga book style. However, since characters usually have their unique signatures, such as hairstyles and clothing, we can integrate the results of the character's face and body to include more character identifiers, so as to compensate for similarities in painting styles. 
These two conjectures enlighten us that making good use of manga content-related features, such as temporal relationship and face-body combination, can help us overcome the difficulties of manga artistic expressions and drawing limitations to achieve good clustering results of manga characters.

Based on the above insights, our contribution includes the following three aspects:

\begin{itemize}
    \item To the best of our knowledge, we are the first to propose and study the unsupervised manga character re-identification (UManga-ReID) task. We propose a Face-body and Spatial-temporal Associated Clustering (FSAC) framework for UManga-ReID. 
    \item We design a temporal-spatial-related loss to fine-tune the pseudo-labeling of manga characters, so as to weaken the challenges posed by artistic expressions. We take advantage of the characters' unique signatures and combine the face image with the body image of the same character for re-identification to overcome the effects of drawing limitations, etc.
    \item Experimental results show that our approach achieves superior performance on both face, body and mixed manga image datasets on Manga109, outperforming the state-of-the-art by a large margin.
\end{itemize}


\begin{table}[t]
\centering
\caption{The proportion of characters that appear more than once in $k$-nearest frames. The data reveals that most of the characters tend to appear in neighboring frames.}
\begin{tabular}{|c|c|c|c|}
\hline
 & \textbf{$k$=1} &\textbf{$k$=3} & \textbf{$k$=5} \\ \hline
\textbf{proportion}       &    70.26\%          & 90.39\%      & 94.77\%  \\ \hline   
\end{tabular}\label{tab:proportion}
\end{table}

\begin{figure*}[t]
\centering\includegraphics[scale=0.4]{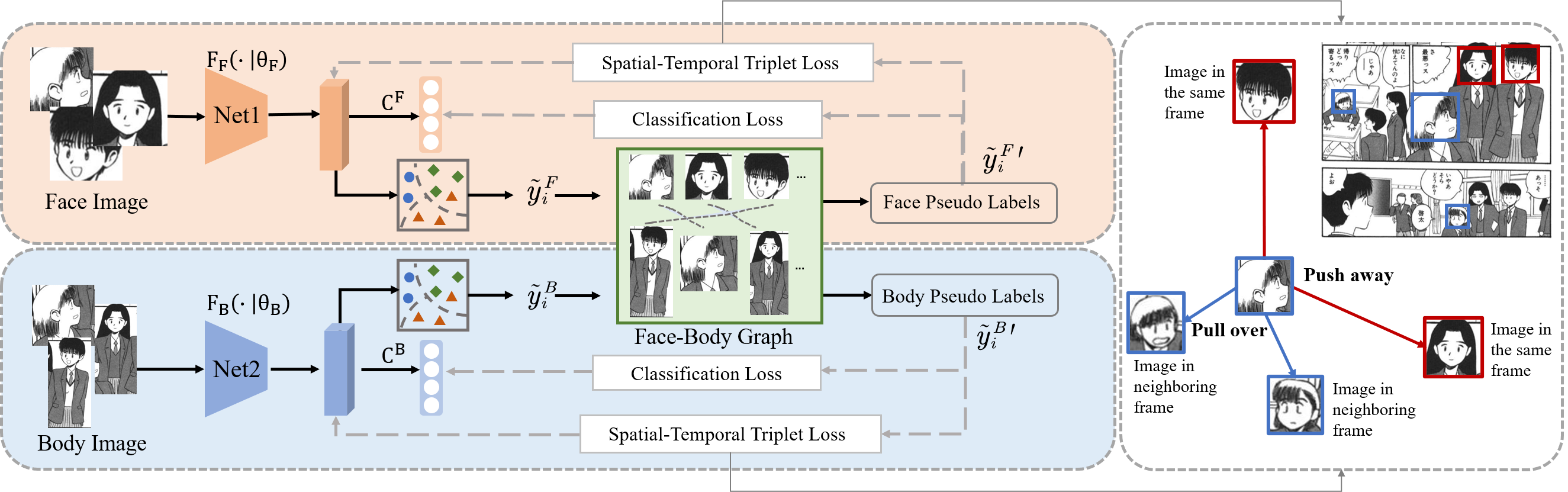}
\caption{The framework of our proposed method with face-body combination and spatial-temporal correction for clustering. The face-body graph combines the predictions of face and body images of the same character to provide comprehensive pseudo labels. The spatial-temporal triplet loss pulls over images in neighboring frames and pushes away images in the same frame.}\label{fig:framework}
\end{figure*}

\section{Related Works}
In line with the focus of our work, we briefly review previous works in the following areas: 1) manga-related works, 2) unsupervised person re-identification methods, and 3) manga character identification works.
\subsection{Manga-related works}
Manga-related researches have increased with the popularity of e-manga.
Some studies such as \cite{aaaimangagan} focus on comic images,  managing to generate comic-style images for real people using GAN.
Some articles concentrate on manga translation. For example, \cite{aaaitrans} designs a fully automated manga translation system via a multi-modal and context-aware translation framework.
There are also studies focusing on the content of comics. \cite{amazing} analyzes the closure-driven narratives of the comics, and \cite{aaaiframe1} and \cite{aaaiframe2} focus on the extraction and order of manga frames(or storyboards).
Some of the well-known cartoon-related datasets include Manga109 \cite{manga1092}, Danbooru \cite{danbooru}, iCartoonFace \cite{cartoon}, etc. Manga109 provides 109 manga books with marks of the face and body images of characters, manga frames and texts, which provides a basis for us to study the clustering of manga characters from the perspective of the spatial-temporal relationship and face-body combination.

\subsection{Unsupervised person re-identification}
Prior to our focus on the comics world,  retrieval of the pedestrian images in the real world had received a lot of attention. 
Person re-identification aims to retrieve the image of a specific pedestrian across the camera and is widely used in surveillance scenes. 

Among many deep learning methods, unsupervised domain adaptation (UDA) methods have attracted much attention because they do not rely on manual annotation. 
\cite{udafan,udalin} provide a baseline of this approach. These methods usually consist of three steps: 1) They extract features of images in the target domain using models pre-trained in the source domain, 2) the generated features are used for clustering to generate pseudo-labels, and 3) pseudo-labels are used to further fine-tune the model. 
However, this method produces hard labels with noise. How to solve the noise problem is the key.
On this basis, \cite{MMT} designs a mutual mean-teaching framework and achieved effective results in unsupervised domain adaptive problems with open sets (i.e., the number of classes of images in the target domain is unknown).

In addition, problems such as image occlusion, appearance change, and pose transformation also plague person re-identification research.

Various methods have been proposed to cope with the appearance change problem in person re-identification research.
\cite{huang2019dot} designs a Domain-Transferred Graph Neural Network to obtain robust representation for the group image. The proposed transferred graph addressed the challenge of the appearance change, while the graph representation overcomes the challenge of the layout and membership change. \cite{xue2018clothing} proposed a cloth-Clothing Change Aware Network (CCAN) and addressed the appearance change issue by separately extracting the face and body context representation. And similar idea is applied in \cite{wan2020person}.

Some patch-based methods learn the partial/regional aggregation properties to make them robust against pose transformation and occlusion. \cite{patchpul} segmented the human body through a patch generation network and developed a PatchNet to learn discriminative features from patches instead of the whole images. \cite{patchssg} proposed a self-similarity grouping (SSG) approach, which exploits the potential similarity (from the global body to local parts) of unlabeled samples to build multiple clusters from different views automatically.
These methods of combining the part and the whole have achieved effective results, and also provide ideas for us to migrate in the manga.

\subsection{Manga character identification}
To date, a limited amount of published literature has reported research on manga character identification. The existing research can be broadly classified into supervised and unsupervised methods.
Some supervision methods, such as \cite{cartoon}, draw on the experience of face recognition methods and propose a method that uses domain data of real-world faces to help classify hyperplanes in the manga domain. This supervised approach can achieve satisfactory performance, but it requires a lot of labeling work, which brings limitations in the manga context.

Unsupervised methods such as \cite{sketch} and \cite{sketch2} proposed a multiscale histogram of edge directions for sketch-based manga retrieval. However, these methods do not consider the diversity among individual manga volumes.  
\cite{adaptation} improves the clustering performance by adapting the manga face representation to the target volume, while manual rules are used in constructing pseudo-label pairs, which lack some generality.



\section{The Proposed Formulation}
\label{sec:method}
In this section, we first introduce the preliminaries and revisit the generic clustering-based unsupervised domain adaptation (UDA) process in Sec. \ref{sec:preliminary}. Then we present the proposed Face-body and Spatial-temporal Associated Clustering (FSAC) method in Sec. \ref{sec:proposedmethod}. Further, we introduce two key modules in Sec. \ref{sec:facebody} and Sec. \ref{sec:spatialtemporal}: the face-body combination module and the spatial-temporal relationship correction module, respectively.

\subsection{Preliminary}
\label{sec:preliminary}
In unsupervised re-identification based on clustering,
given a target training set $X^t=\{x_1^t,x_2^t,...,x_{N^t}^t\}$ of $N^t$ images, the goal is to learn a feature embedding function $F(\theta;x_i^t)$ from $X^t$ without any manual annotation, where parameters of $F$ are collectively denoted as $\theta$.

For UDA methods, however, an additional source dataset $X^s=\{x_1^s,x_2^s,...,x_{N^s}^s\}$ of $N^s$ images is typically used for pre-training, aiming at learning domain-invariant features by both the source and target domains.
The UDA approach applies the classifier $F(\cdot|\theta)$ learned from the source domain data to the target domain data with no or little labeling in the target domain. The source domain images' and target domain images’ features encoded by the network are denoted as
$\left\{ F\left( x_{i}^{s}|\theta \right) \right\} |_{i=1}^{N_s}$ and $\left\{ F\left( x_{i}^{t}|\theta \right) \right\} |_{i=1}^{N_t}$, respectively.

To learn the feature embedding, traditional clustering-based UDA methods usually consist of three steps: 

1) Features $\left\{ F\left( x_{i}^{t}|\theta \right) \right\} |_{i=1}^{N_t}$ of images in the target domain are extracted using the model $F(\cdot|\theta)$ pre-trained in the source domain;

2) By using the generated features $\left\{ F\left( x_{i}^{t}|\theta \right) \right\} |_{i=1}^{N_t}$, $x_i^t$ images are clustered into $M^t$ classes.  Let $\tilde{y}_{i}^{t}$ denote the pseudo label generated for image $x_t$; 

3) Using the generated pseudo-labels, the model parameters $\theta$ and a learnable classifier $C^t:\rightarrow f^t\left\{ 1,\cdots ,M_t \right\}$ of the target domain dataset are fine-tuned by the classification loss and triplet loss denoted as:

\begin{equation}
\label{eq:udacls}
\begin{aligned}
\mathcal{L} _{id}^{t}\left( \theta \right) =\frac{1}{N_t}\sum_{i=1}^{N_t}{\begin{array}{c}
    \mathcal{L} _{ce}\left( C^t\left( F\left( x_{i}^{t}|\theta \right) \right) ,\tilde{y}_{i}^{t} \right)\\
\end{array}}
\end{aligned}
\end{equation}

\begin{equation}
\label{eq:udatri}
\begin{aligned}
    \mathcal{L}_{tri}^{t}(\theta)  
    & =\frac{1}{N_t}\sum_{i=1}^{N_t} \max ( 0,\| F( x_i^t|\theta) - F( x_{i,p}^t|\theta) \|  +m \\
    & - \| F( x_i^t|\theta) -F( x_{i,n}^t|\theta) \| ),
\end{aligned}
\end{equation}
where $\left\| \cdot \right\|$ denotes $L2$-norm distance, subscripts ${i,p}$ and ${i,n}$ represent the positive and negative feature index in each mini-batch for the sample $x_i^t$, and $m$ denotes the triplet distance margin. 

This feature embedding function can be applied to the target gallery set, $X_g^t=\{x_{g_1}^t,x_{g_2}^t,...x_{g_{N_g}^t}\}$ of $N_g^t$ images, and the target query set $X_q^t=\{x_{q_1}^t,x_{q_2}^t,...x_{q_{N_q}^t}\}$ of $N_q^t$ images. During the evaluation, we use the feature of a target query image $F(\theta;x_{q_i}^t)$ to search similar image features from the target gallery set. The query result is a ranking list of all gallery images according to the Euclidean distance between the feature embedding of the target query and gallery data, {\it i.e.},
\begin{equation}
    d({x^t_q}_i,{x^t_g}_i))=\left\| F(\theta ;{x^t_q}_i) -F(\theta ;{x^t_g}_i) \right\|.
\end{equation}
The feature embeddings are supposed to assign a higher rank to similar images and keep the images of a different person a low rank.

\subsection{Our proposed method}
\label{sec:proposedmethod}
In this section, we present the proposed Face-body and Spatial-temporal Associated Clustering (FSAC) method. 
As mentioned above, manga characters have special features, such as spatial-temporal relationships and unique signatures of the characters, which may help clustering.

Therefore, our core idea is to make use of the two special features to make up for the shortage of clustering.
To achieve this goal, we propose two modules: the face-body combination module and the spatial-temporal relationship correction module.

Overall, the framework of our method is shown in Figure~\ref{fig:framework}. The target dataset is divided into related face-set and body-set, where each face image is derived from a crop of body images and they are one-to-one correspondence.
Formally, we denote the body data as $\mathbb{D}_B=(x_{i}^{B},k_{i}^{B})|_{i=1}^{N_B}$, 
where $x_{i}^{B}$ denotes the $i$-th body training sample, $k_{i}^{B}$ is the data associated with the spatial-temporal information of $x_{i}^{B}$, indicating the manga frame index of the image $x_{i}^{B}$, and $N_B$ stands for the number of body images. We also denote the face data as $\mathbb{D}_F=(x_{i}^{F},k_{i}^{F})|_{i=1}^{N_F}$, where $x_{i}^{F}$ and $k_{i}^{F}$ denote the $i$-th face training sample and the manga frame index of the training image, and $N_F$ stands for the number of face images. 

For each part, we first extract features which are denoted as $
\left\{ F_B\left( x_{i}^{B}|\theta_B \right) \right\} |_{i}^{N_B}
$
and $
\left\{ F_F\left( x_{i}^{F}|\theta_F \right) \right\} |_{i}^{N_F}
$
through their networks, respectively. 
Then, we employ a clustering algorithm on the unlabeled data and the pseudo labels $\tilde{y}_i^{B'}$ and $\tilde{y}_{i}^{F'}$ are generated under the constraint of a face-body graph. 
After that, based on these pseudo labels, we optimize the network parameters $\theta$ and learnable classifier $C^t: f^t\rightarrow \left\{ 1,\cdots ,M^t \right\}$ jointly by the classification loss and the spatial-temporal triplet loss described in Equation~\ref{eq:cls} and Equation~\ref{eq:tri}.
The above steps are repeated until convergence.

\subsection{The Face-body Combination Module}
\label{sec:facebody}
In the beginning of clustering, each image is assigned to a different cluster for initialization, and then similar images are gradually clustered together by distance.
The cluster ID is used as the training set label, and we expect the network to minimize the intra-cluster variance and maximize the inter-cluster variance.

Following \cite{soft1,soft2,soft3}, we generate soft pseudo labels for face and body after clustering, which are expressed in terms of probability. 
For each image, the probability that it belongs to the $m$-th class is defined as
\begin{equation}
\label{eq:soft}
\tilde{y}^m=p\left( y_m|x,\{a_i\}_{i=1}^{N} \right) =\frac{\exp \left( a_{m}^{\top}F\left( x|\theta \right) \right)}{\sum_{i=1}^N{\exp \left( a_{i}^{\top}F\left( x|\theta \right) \right)}},
\end{equation}
where ${\{a_i\}}_{i=1}^N$ is the reference agent that stores the feature of each class and $N$ is the number of classes.

The core component of the face-body combination module is the face-body graph, which builds a map of the body and face images of the same character from the same manga frame.
For face images $x_i^F$ and body images $x_i^B$ with mapping relationship, their soft labels are respectively expressed as $\tilde{y}_{i}^{F}$ and $\tilde{y}_{i}^{B}$. 
The reference agents corresponding to the soft label with the maximum probability are denoted as $m_i^F$ and $m_i^B$.
The graph enables us to synthesize the face and body results in the following way:

\begin{equation}
\label{eq:combine}
\tilde{y}_{i}^{B'}=\tilde{y}_{i}^{F'}=
\left\{ 
\begin{aligned}
    m_i^B, \tilde{y}_i^B   >    \tilde{y}_i^F \\
    m_i^F, \tilde{y}_i^B  \leq  \tilde{y}_i^F   .
\end{aligned}
\right.
\end{equation}

The refined labels $\tilde{y}_{i}^{B'}$ and $\tilde{y}_{i}^{F'}$ will be further used in the fine-tune network.

\subsection{The Spatial-temporal Relationship Correction Module}
\label{sec:spatialtemporal}
In this module, we add spatial-temporal constraints on the basis of the classic triplet loss \cite{triplet} and optimize network parameters jointly with the classification loss ({\it i.e.}, the cross-entropy loss).


The classic triplet loss and classification loss are shown in Equation \ref{eq:udacls} and Equation \ref{eq:udatri}. In the traditional triplet loss, the positive and negative sample pairs are selected based on the feature distance from the anchor pictures.
However, in the UManga-ReID task, we find that spatial-temporal information may help clustering, so we set up spatial-temporal correlation rules to help select positive and negative sample pairs.

In order to add spatial-temporal constraints on the triplet loss, we define a spatial-temporal distance $d^{st}$ to find the positive and negative pairs of the sample $x_i$. The spatial-temporal distance between $x_i$ and $x_j$ is denoted as
\begin{equation}
d_{i,j}^{st}=\min \left( \sigma ,\left| k_i-k_j \right| \right) +\eta,
\label{eq:dstij}
\end{equation}
where $\sigma$ indicates the threshold of manga frame index difference and $\eta$ denotes the penalty factor of characters in the same manga frame in the form of
\begin{equation}
    \eta =
\begin{cases}
    0, k_i\ne k_j \\
    \eta, k_i=k_j .\\
\end{cases}
\end{equation}

The sum of the spatial-temporal distance and the $L2$-norm distance of images constitutes the total distance $d_{i,j}$, denoted as
\begin{equation}
\label{d}
d_{i,j}=\left\| F\left( x_i\left| \theta \right. \right) -F\left( x_j\left| \theta \right. \right) \right\| +d_{i,j}^{st},
\end{equation}
which is used to search for pseudo positive and negative pairs in the triplet loss.

Meanwhile, we still use the $L2$-norm distance of the pseudo positive and negative pairs to calculate the triplet loss.

Taking the face dataset as an example, the spatial-temporal triplet loss and classification loss shown below are are still calculated in the classical way, but differ in some details.

\begin{equation}
\label{eq:cls}
\begin{aligned}
\mathcal{L} _{id}^{F}\left( \theta \right) =\frac{1}{N_F}\sum_{i=1}^{N_F}{\begin{array}{c}
    \mathcal{L} _{ce}\left( C^F\left( F\left( x_{i}^{F}|\theta \right) \right) ,\tilde{y}_{i}^{F} \right)\\
\end{array}}
\end{aligned}
\end{equation}

\begin{equation}
\label{eq:tri}
\begin{aligned}
    \mathcal{L}_{st-tri}^{F}(\theta)  
    & =\frac{1}{N_F}\sum_{i=1}^{N_F} \max ( 0,\| F( x_i^F|\theta) - F( x_{i,p}^F|\theta) \|  +m \\
    & - \| F( x_i^F|\theta) -F( x_{i,n}^F|\theta) \| ),
\end{aligned}
\end{equation}

In classification loss, $\tilde{y}_{i}^{F}$ uses the refined label after face-body combination, and in spatial-temporal triplet loss, positive and negative pairs are selected based on spatial-temporal distance $d_{st}$.

\tabcolsep=4pt
\begin{figure*}[!htb]
	\centering
\footnotesize{
		\begin{tabular}{cccc}
			\includegraphics[width=0.2\textwidth]{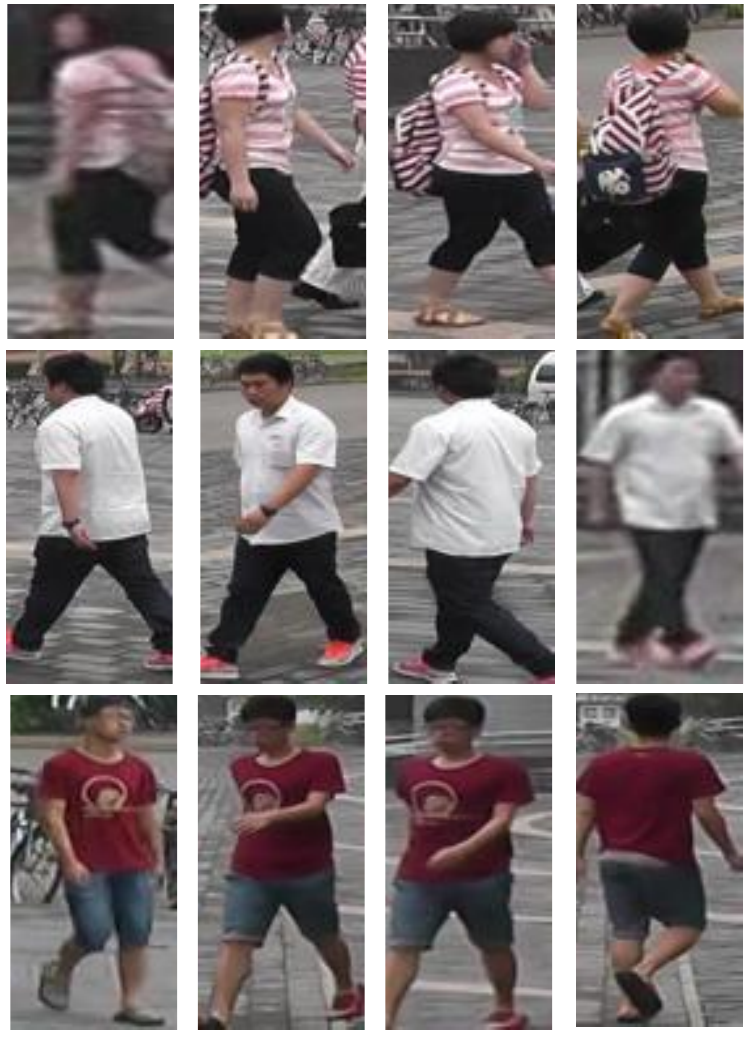} &
			\includegraphics[width=0.29\textwidth]{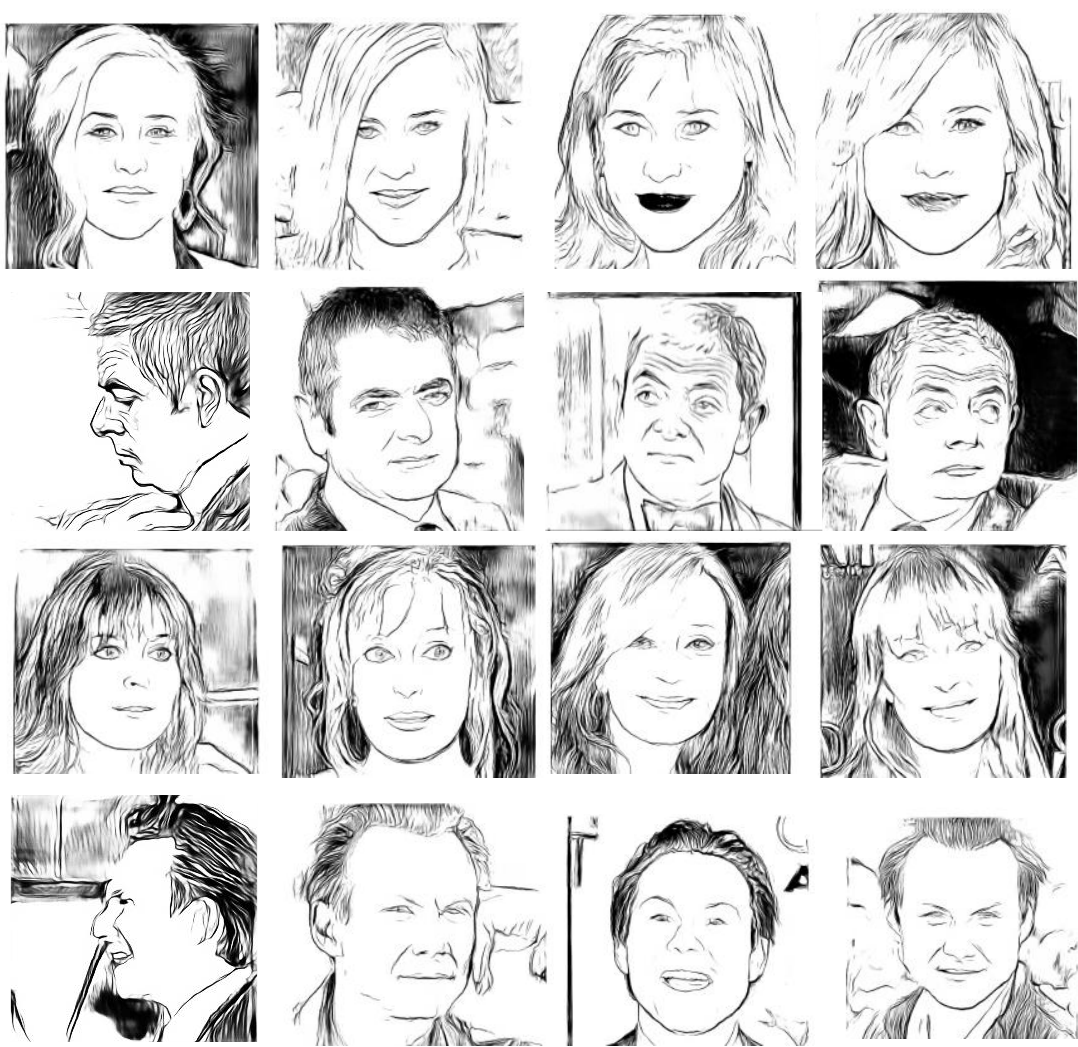} &
			\includegraphics[width=0.235\textwidth]{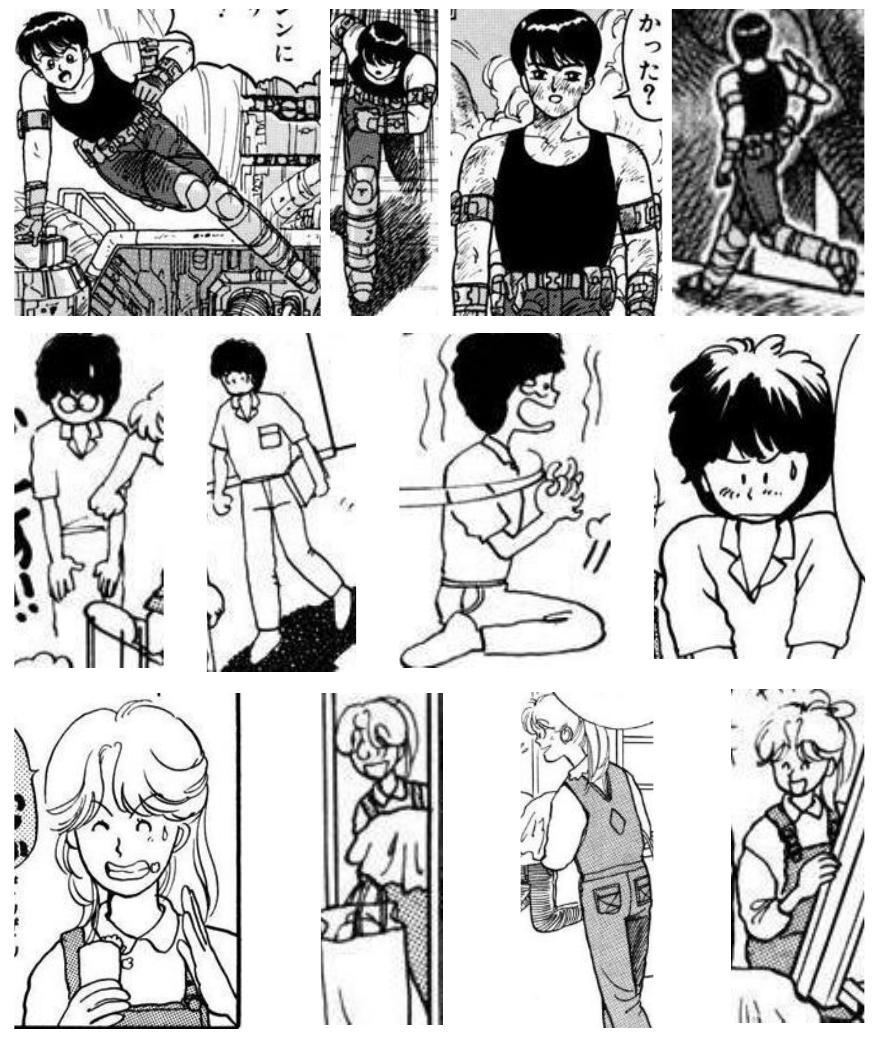} &
			\includegraphics[width=0.252\textwidth]{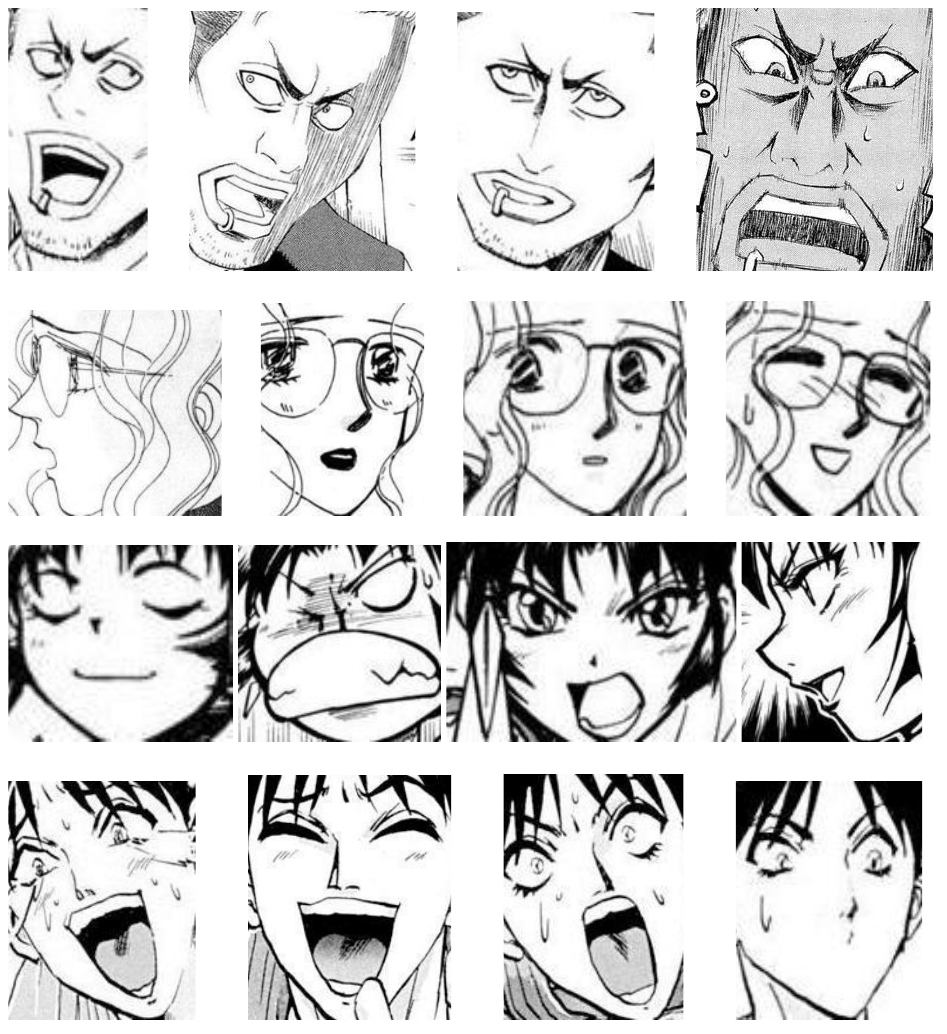} \\
	(a) Source Body & (b) Source Face & (c) Manga Body & (d) Manga Face \\
	\end{tabular}}
\caption{Some examples of used datasets. For each dataset, images in each row are from the same identity. (a) Market-1501~\cite{market1501} is selected as the source body dataset; (b) CelebA~\cite{celeba} sketch styled by u2-net network~\cite{u2net} is selected as the source face dataset; (c) and (d) are the testing manga body and face datasets from Manga109~\cite{sketch,manga1092}.
   }
\label{fig:examples}
\end{figure*}

\begin{table*}[t]
\caption{Comparison with the state-of-the-art methods on Manga109-face and Manga109-body datasets. Baseline is the general UDA pipeline illustrated in Sec.\ref{sec:preliminary} in \cite{MMT}.Strong-baseline is the updated version of baseline.}
\label{tab:compare}
\centering
\tabcolsep=8pt
\begin{tabular}{@{}l|l|llll@{}}
\toprule
Datasets                      & Methods                  & mAP    & rank-1 & rank-5 & rank-10 \\ \midrule
\multirow{5}{*}{Face Dataset} & Direct infer on resnet50 & 7.1\%  & 22.5\% & 39.6\% & 59.1\%  \\
                              & Baseline \cite{MMT} (in Sec. \ref{sec:preliminary})                & 23.9\% & 63.9\% & 78.9\% & 82.9\%  \\
                              & Strong Baseline (update of baseline)          & 24.4\% & 51.7\% & 66.1\% & 70.6\%  \\
                              & MMT \cite{MMT} (in Sec. \ref{sec:preliminary})                      & 21.3\% & 50.6\% & 62.9\% & 71.7\%  \\
                              & Our Method               & \textbf{32.2\%} & \textbf{78.4\% }& \textbf{88.1\%} &\textbf{ 91.7\%}  \\ \midrule
\multirow{5}{*}{Body Dataset} & Direct infer on resnet50 & 0.6\%  & 2.8\%  & 7.4\%  & 10.0\%  \\
                              & Baseline  \cite{MMT} (in Sec. \ref{sec:preliminary})                & 5.9\%  & 32.0\% & 49.0\% & 57.1\%  \\
                              & Strong Baseline (update of baseline)         & 5.2\%  & 28.8\% & 37.2\% & 46.0\%  \\
                              & MMT \cite{MMT} (in Sec. \ref{sec:preliminary})                      & 4.8\%  & 19.4\% & 30.9\% & 36.3\%  \\
                              & Our Method               & \textbf{10.2\%} &\textbf{ 40.3\%} & \textbf{66.8\%} & \textbf{71.1\%}  \\ \midrule
\multirow{5}{*}{Face Dataset + Body Dataset} & Direct infer on resnet50 & 0.5\%  & 16.9\% & 21.1\% & 23.5\%  \\
                              & Baseline \cite{MMT} (in Sec. \ref{sec:preliminary})                & 3.2\% & 31.4\% & 45.5\% & 53.7\%  \\
                              & Strong Baseline (update of baseline)          & 2.8\% & 33.1\% & 39.6\% & 51.9\%  \\
                              & MMT \cite{MMT} (in Sec. \ref{sec:preliminary})                      & 4.8\% & 38.8\% & 52.7\% & 60.0\%  \\
                              & Our Method               & \textbf{9.8\%} & \textbf{52.8\% }& \textbf{68.6\%} &\textbf{ 73.1\%}  \\ 
                              \bottomrule
\end{tabular}
\end{table*}

\section{The Proposed Algorithm}


\begin{algorithm}[t]
\caption{The FSAC framework}
\label{alg:1}
{\bf Input:}
Unlabeled manga body data $\mathbb{D}_B=(x_{i}^{B},k_{i}^{B})|_{i=1}^{N}$;\\
Unlabeled manga face data $\mathbb{D}_F=(x_{i}^{F},k_{i}^{F})|_{i=1}^{N}$;\\
$F_B\left( \cdot |\theta _B \right)$ pre-trained on real human body data;\\
$F_F\left( \cdot |\theta _F \right)$ pre-trained on real human face data.\\
{\bf Output:} 
The learned CNN model $F_B\left( \cdot |\theta _B \right)$ and $F_F\left( \cdot |\theta _F \right)$.
\begin{algorithmic}[1]
\State Initialize the frame threshold $\sigma$ and penalty factor $\eta$ for Equation~\ref{eq:dstij}. 
\For{$n$ in $[1,num\_epochs]$} 
\State Extract body feature $\left\{ F\left( x_{i}^{B}|\theta \right) \right\} |_{i}^{N_B}$ and face feature $\left\{ F\left( x_{i}^{F}|\theta \right) \right\} |_{i}^{N_F}$;
\State Generate soft pseudo labels $\tilde{y}_{i}^{B}$ for each sample $x_i^B$ in $\mathbb{D}_B$ and $\tilde{y}_{i}^{F}$ for $x_i^F$ in $\mathbb{D}_F$ following Equation~\ref{eq:soft};
\For{each mini-batch $B$, iteration $T$}
\State Generate combined pseudo labels $\tilde{y}_{i}^{B'}$ and $\tilde{y}_{i}^{F'}$ via face-body graph following Equation \ref{eq:combine}; 
\State Update parameters $\theta_B$ and $\theta_F$ via the spatial-temporal loss and classification loss following Equation \ref{eq:tri} and \ref{eq:cls} with the spatial-temporal distance.
\EndFor
\EndFor
\end{algorithmic}
\end{algorithm}

As demonstrated in Figure~\ref{fig:framework}, our framework consists of two parts: the face part and the body part, which are interconnected but do not interfere with each other.
Our goal is to optimize their respective models by taking advantage of their common relationships and their internal spatial-temporal relationships.

The proposed FSAC algorithm consists of two steps: (1) pre-training phase, which initializes the CNN network by pre-training with the labeled source domain datasets; (2) loop iteration phase, which boosts clustering leveraging the proposed two modules. We elaborate on the two steps as follows.

\subsection{Pre-training stage}

We use the labeled real-person face and body datasets as source domains for pre-training. These two pre-trained models $F_F\left( \cdot |\theta _F \right)$ and $F_B\left( \cdot |\theta _B \right)$ will be used as the CNN initialization for the face and body in the subsequent loop iteration steps, respectively.

\subsection{Loop iteration stage}
In the iterative process, we first extract image features and generate soft pseudo-labels after clustering.
Then, following Equation \ref{eq:combine} in the face-body combination module, we generate the same hard pseudo-label $\tilde{y}_{i}^{B'}=\tilde{y}_{i}^{F'}$ for the corresponding face and body images.
After that, we employ the spatial-temporal triplet loss and the classification loss to update their respective networks.

Such operations of generating pseudo labels by clustering and learning features with pseudo labels are alternated until the training converges. 
The entire process is summarized in Algorithm \ref{alg:1}.

\section{Experiments}
\subsection{Datasets}

The experimental dataset includes four parts: the source face dataset and source body dataset for pre-training face and body networks, as well as the target face dataset and target body dataset for fine-tuning on the manga domain. The details are as follows.

\subsubsection{Market-1501 (source body dataset)}
Market-1501\cite{market1501} is a large-scale public benchmark dataset for person  re-identification. It contains 1501 identities which are captured by six different cameras,  and 32,668 pedestrian image bounding-boxes obtained using the deformable part models pedestrian detector. Each person has 3.6 images on average at each viewpoint. The dataset is split into two parts: 750 identities are utilized for training and the remaining 751  identities are used for testing. In the official testing protocol, 3,368 query images are selected as probe set to find the correct match across 19,732 reference gallery images. The division of data sets and specific data are shown in table \ref{tab:dataset}.

\subsubsection{CelebA-U2Net (source face dataset)}
CelebFaces Attributes Dataset (CelebA) \cite{celeba} is a large-scale face attributes dataset with more than 200K celebrity images, each with 40 attribute annotations. The images in this dataset cover large pose variations and background clutter. CelebA has large diversities, large quantities, and rich annotations, including 10,177 identities, 202,599 number of face images, and 5 landmark locations, 40 binary attributes annotations per image.

To achieve better performance in the manga context, we use U2-Net \cite{u2net} to acquire a sketch-style human face dataset, which is called CelebA-U2Net. 
U2-net is a simple yet powerful deep network architecture for salient object detection and provides an interesting application for human portrait drawing, which helps us transform real-face pictures into sketch-style pictures. This helps us to have more commonalities between the source dataset and target dataset, which is conducive to subsequent training and learning.
In the process of style migration, we deleted images with undetectable faces. The new-obtained dataset is described in Table \ref{tab:dataset}.

\subsubsection{Manga109-body (target body dataset)}
Manga109 \cite{sketch,manga1092} is a dataset of a variety of 109 Japanese comic books publicly available for use for academic purposes. Frames, texts, and the face and body images of characters are defined and labeled in each manga book.


To improve the applicability of the network, we select the characters that appear more than 9 times for face and body data respectively.
We find and match face and body images of the same character in the same manga frame, and build face and body datasets that map to each other. Generally, the face image is the face part of its mapped body image and the face dataset and body dataset share the same image number and character number.

After that, we have a face dataset and a body dataset with 186,817 images and 1506 characters each.
We divided the datasets into training sets and gallery sets in a ratio of 2:1 and randomly selected one image from the gallery for each character as their respective query set. Therefore, the query set of face and body data each has 502 images of different characters.
Face images and body images are resized to 112 × 112 and 128 × 256 correspondingly before being fed into the networks.

\subsubsection{Manga109-face (target face dataset)}
Manga109-face takes the same processing as manga109-body but uses a subset of the faces. Relevant statistics are shown in Table \ref{tab:dataset} and sample images of the datasets can be found in Figure \ref{fig:examples}.

\begin{table}[]
\centering
\caption{Statistics of datasets.}
\label{tab:dataset}
\begin{tabular}{@{}c|c|ccc@{}}
\toprule
\textbf{dataset}               & \textbf{size}            & \textbf{subdataset} & \textbf{images} & \textbf{identities} \\ \midrule
\multirow{3}{*}{Market-1501}   & \multirow{3}{*}{64*128}  & train               & 12937           & 750                 \\
                               &                          & query               & 3368            & 751                 \\
                               &                          & gallery             & 19732           & 751                 \\ \midrule
CelebA-U2Net                   & 112*112                  & \textbackslash{}    & 194098          & 10172               \\ \midrule
\multirow{4}{*}{Manga109-Body} & \multirow{4}{*}{64*128}  & train               & 137482          & 1004                \\
                               &                          & query               & 502             & 502                 \\
                               &                          & gallery             & 48833           & 502                 \\
                               &                          & total               & 186817          & 1506                \\ \midrule
\multirow{4}{*}{Manga109-Face} & \multirow{4}{*}{112*112} & train               & 137482          & 1004                \\
                               &                          & query               & 502             & 502                 \\
                               &                          & gallery             & 48833           & 502                 \\
                               &                          & total               & 186817          & 1506                \\ \bottomrule
\end{tabular}
\end{table}



\subsection{Implementation Details}
\subsubsection{On pre-training stage}
We use ResNet50 \cite{resnet} pre-trained on real-human data for the convolution layers.
Considering the huge differences between face and body images, we used different pre-trained models for them respectively.

Specifically, for the body part, we used the real-world pedestrian dataset Market-1501~\cite{market1501} for pre-training. Market-1501 is the currently commonly used pedestrian re-recognition dataset, consisting of 12,937 training images (from 750 different people), and 19,732 test images (from another 751 different people). We pre-train the ResNet50 model on Market-1501 via the MMT method \cite{MMT} and select one of its models as the body pre-trained model. 

For the face part, the CelebA~\cite{celeba} dataset is selected while we find that sketch-style human images improve the clustering performance in the manga context. Therefore, we use the u2-net~\cite{u2net} to transfer the CelebA images into black and white sketch style for pre-training.
Unlike the body part, ArcFace~\cite{arcface} is used as the classifier for face data pre-training, which is validated to be a better way of mining the features of faces.
The pre-training of face and body networks are initialized with ImageNet~\cite{imagenet} pre-trained weights. Figure~\ref{fig:examples}(a) and \ref{fig:examples}(b) show some samples of Market-1501 and CelebA.

\subsubsection{On loop iteration stage}
Based on the pre-trained models, we then fine-tune the network through face-body and spatial-temporal modules.
On the clustering stage, we use DBSCAN as the clustering algorithm (k-means is also supported given the number of clustering classes). In the calculation of spatial-temporal distance $d^{st}$, we set the threshold of manga frame index difference $\sigma =100$ and penalty factor $\eta = 1000$ in Equation \ref{eq:dstij}.
In order to make full use of the spatial-temporal relationship, a no-shuffle method was selected during the sampling.

\subsubsection{Evaluation metrics}
We employ the Mean Average Precision (mAP) and Cumulative Matching Characteristics \cite{cmc} for evaluation.
The mAP reflects the recall, while CMC scores reflect the retrieval precision.
We report the rank-1, rank-5 and rank-10 scores to represent the CMC curve.

\begin{figure}[t]
\centering\includegraphics[scale=0.45]{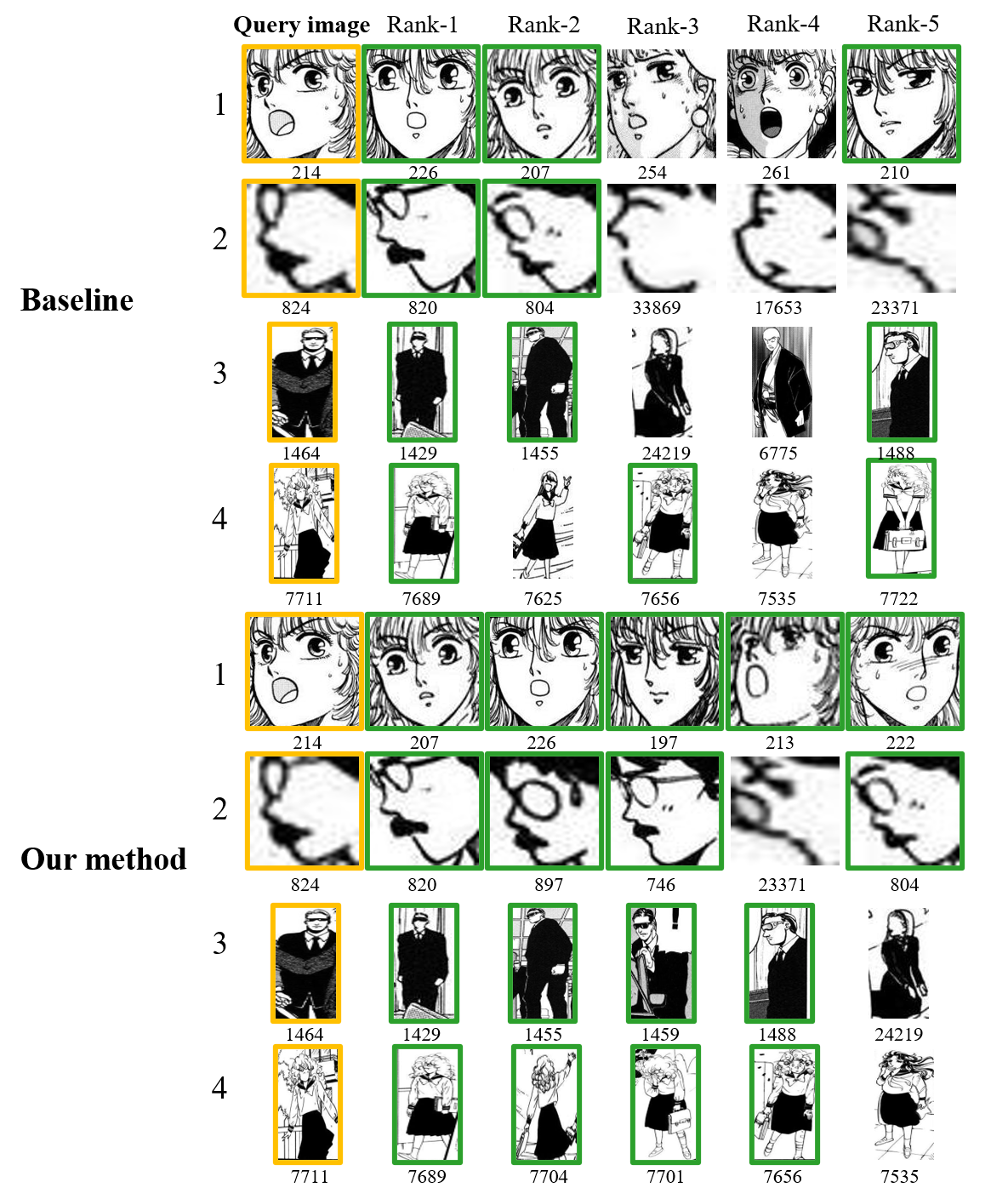}
\caption{Query results. The first image is the search image, followed by the rank-1 to rank-5 results. The images framed in green are the correct results. The numbers below the images denote the frame sequence index of the image. The difference between the two frame sequence index reveals the temporal distance of the two images. }\label{fig:query}
\end{figure}

\subsection{Comparison with the State-of-the-Art}
The comparisons with the state-of-the-art methods on Manga109 are shown in Table \ref{tab:compare}.

The baseline method in the table uses the general Unsupervised Domain Adaptation(UDA) pipeline illustrated in Sec. \ref{sec:preliminary} in \cite{MMT}, which adopts features extracted from the pre-trained network to cluster and uses the classification loss and classic triplet loss to update the network, excluding modules for face-body combination and temporal \& spatial relationship modification.
Strong-baseline is the follow-up update of this method, changing the steps of pre-training to synchronous training of shared weights.

Although these methods achieve outstanding performance in person re-identification, our manga-content-based method surpasses them by a large margin in the manga context in terms of mAP and CMC.
On the face dataset, we achieve the best results with mAP of 32.2\% and rank-1 score of 78.4\%, which outperforms the state-of-the-art result by 7.8\% and 14.5\% respectively. On the body dataset, we also surpass the state-of-the-art result by 4.3\% and 8.3\% on mAP and rank-1 score respectively.
When tested with a mixed dataset of Manga109Face and Manga109Body, we achieve 5.0\% and 14.0\% of improvement in rank-1 accuracy and mAP, respectively.

In addition, although \cite{adaptation} also studies this topic, their results were not compared because the dataset is pre-screened and we have no access to their code and experimental data.

\subsection{Ablation study}
We conduct ablation studies on the two key parts: the spatial-temporal triplet loss and the face-body graph. 
For the model without the spatial-temporal triplet loss, the classic triplet loss \cite{triplet} is used. 
As shown in Table~\ref{tab:ablation}, the comparison results on the two datasets validate the effectiveness of the spatial-temporal module, which can be seen from the mAP difference of 5.2\% on the face dataset and 2.2\% on the body dataset.

In addition, for the model without the face-body graph, we adopt the hard pseudo labels generated by clustering instead. The results in Table~\ref{tab:ablation}  also show the effectiveness of the face-body combination module.

\begin{table*}[t]
\centering
\caption{Ablation studies. The model without spatial-temporal triplet loss uses classic triplet loss \cite{triplet} instead. The model without the face-body graph uses the hard label generated by clustering instead.}\label{tab:ablation}
\tabcolsep=8pt
\begin{tabular}{c|c|c|cccc}
\hline
\textbf{Datasets}             & \textbf{Spatial-temporal triplet loss} & \textbf{Face-body Graph} & \textbf{mAP} & \textbf{rank-1} & \textbf{rank-5} & \textbf{rank-10} \\ \hline
\multirow{3}{*}{Face Dataset} &                                   &                          & 23.9\%       & 63.9\%          & 78.9\%          & 82.9\%           \\
                              & \checkmark                                &                          & 29.1\%       & 72.5\%          & 85.5\%          & 89.4\%           \\
                              & \checkmark                                 & \checkmark                       & 32.2\%       & 78.4\%          & 88.1\%          & 91.7\%           \\ \hline
\multirow{3}{*}{Body Dataset} &                                   &                          & 5.9\%        & 32.0\%          & 49.0\%          & 57.1\%           \\
                              & \checkmark                                 &                          & 8.1\%        & 38.8\%          & 60.2\%          & 65.3\%           \\
                              & \checkmark                                 & \checkmark                        & 10.2\%       & 40.3\%          & 66.8\%          & 71.1\%           \\ \hline
\multirow{3}{*}{Face + Body Dataset} &                                   &                          & 3.2\%        & 31.4\%          & 45.5\%          & 53.7\%           \\
                              & \checkmark                                 &                          & 9.3\%        & 50.3\%          & 62.6\%          & 68.8\%           \\
                              & \checkmark                                 & \checkmark                        & 9.8\%       & 52.8\%          & 68.6\%          & 73.1\%           \\ \hline
\end{tabular}
\end{table*}

\begin{figure}[t]
\centering\includegraphics[scale=0.3]{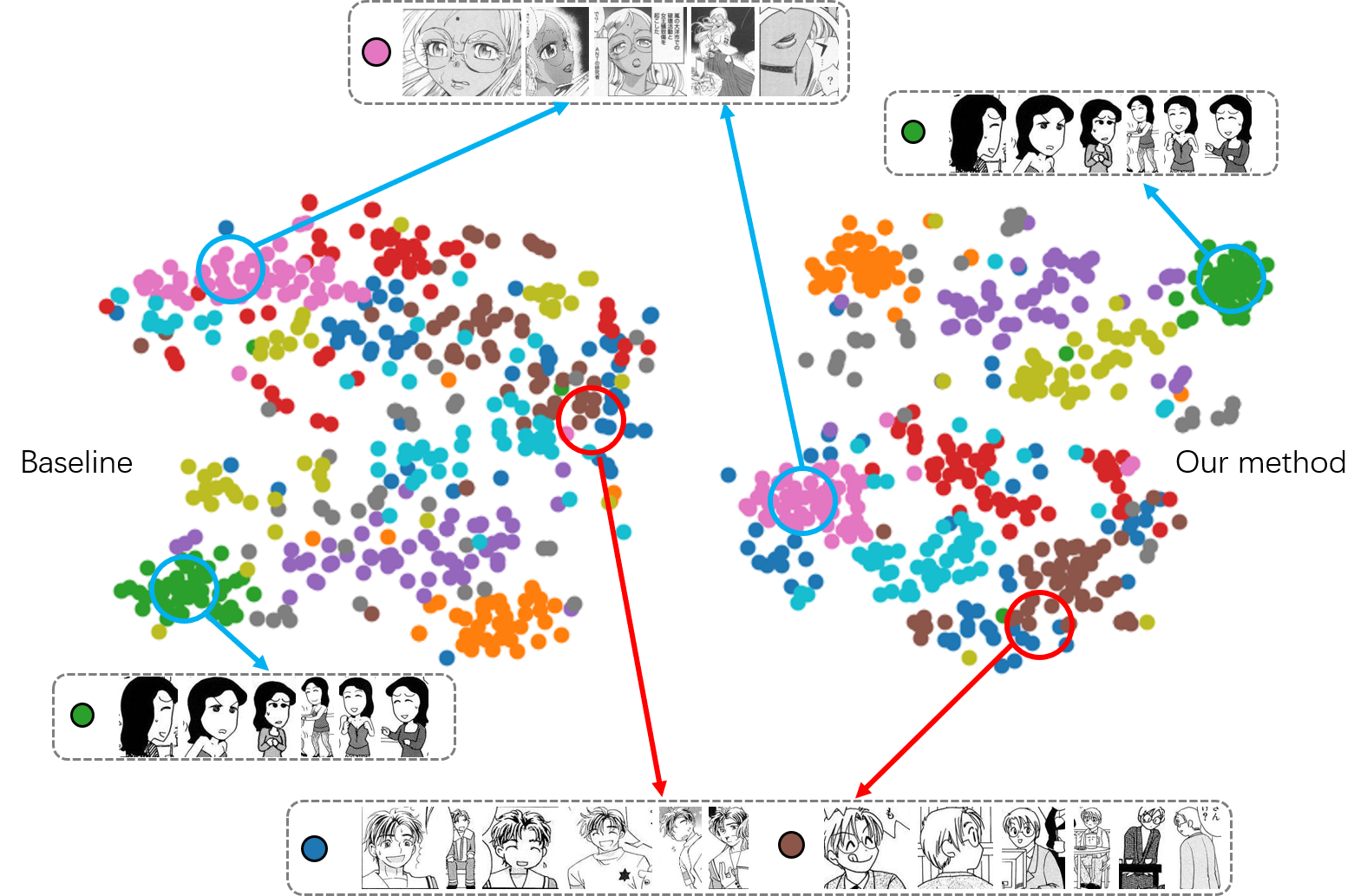}
\caption{Comparison of t-SNE results of our method with baseline method of the learned feature embeddings on a part of the Manga109-body training set (10 characters, 481 images). Points of the same color represent images of the same character. We show detailed images of two positive examples (the blue circle) and a negative example (the red circle). The points in the blue circles have the same identity. In the red circles, the blue and brown points are clustered together, indicating that our algorithm incorrectly combines them into one cluster. However, these samples look very similar and are hard to be distinguished from each other.}\label{fig:cluster}
\end{figure}

\subsection{Algorithmic Analysis}

To make the experimental analysis more comprehensive and to explore the details of the effects of the spatial-temporal relationship, we conducted experimental tests on the parameters $\sigma$ and $\eta$ in Equation \ref{eq:dstij}. The experimental results are displayed in Figure \ref{fig:parameters}.

\subsubsection{Analysis of the parameters}

\textbf{The parameter $\sigma$} is the threshold of the manga frame distance, which intuitively determines the upper limit of the spatial-temporal distance between two images that are not in the same manga frame, and controls to some extent the weight of the manga spatial-temporal distance to the image similarity distance. The results show the performance for $\sigma$ ranging between 50 and 150 in Figure \ref{fig:parameters}(a). Usually, a manga chapter contains 100 to 200 image frames, and the $\sigma$ range taken for our tests was thus determined.
We observed that with the increase of $\sigma$, the experimental results showed a weak trend of decreasing before increasing, but the mAP values are stable around 32.5\%, and the overall values don't fluctuate much.

\textbf{The parameter $\eta$} is a penalty term for two images that are in the same frame. We tested the performance of $\eta$ when taking different orders of magnitude values, and the experimental results are shown in Figure \ref{fig:parameters}(b). The results show that with the increase of $\eta$, the model performance shows a trend of increasing and then decreasing, and the model has better performance when $\eta$ takes the value of about 1000.

\subsubsection{Analysis of the shuffle and no-shuffle training strategy}

\begin{figure*}[!htb]
	\centering
\footnotesize{
		\begin{tabular}{cccc}
			\includegraphics[width=0.24\textwidth]{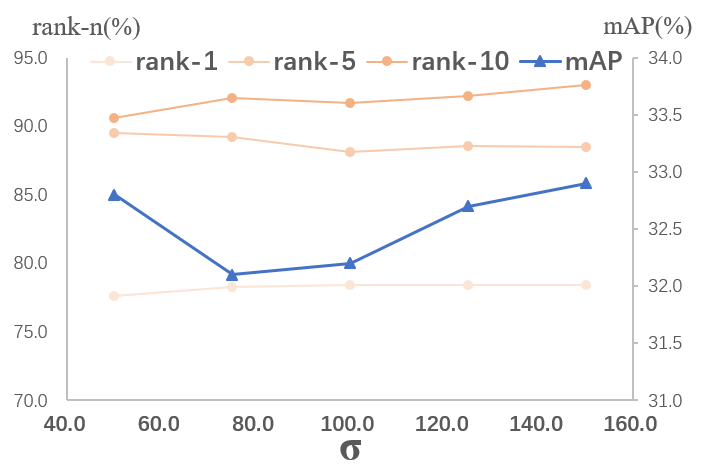} &
			\includegraphics[width=0.24\textwidth]{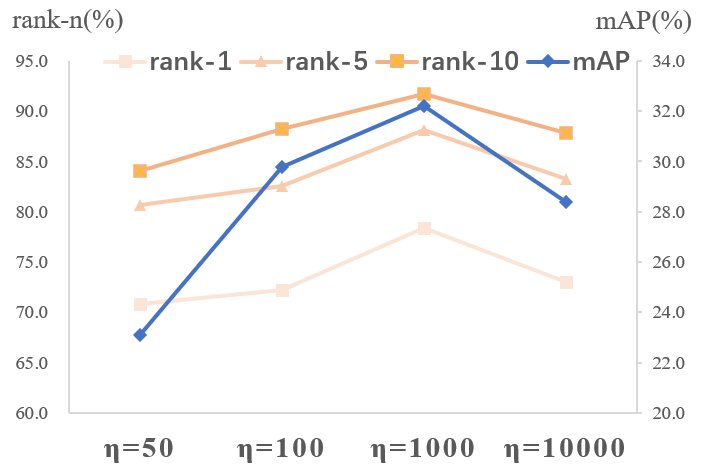} &
			\includegraphics[width=0.24\textwidth]{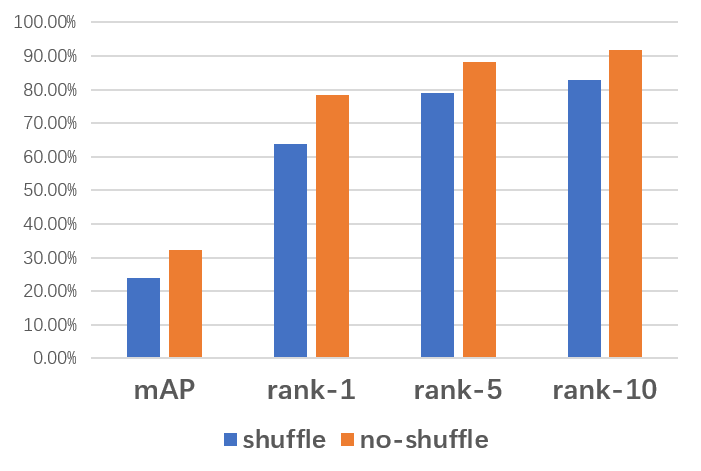} &
			\includegraphics[width=0.24\textwidth]{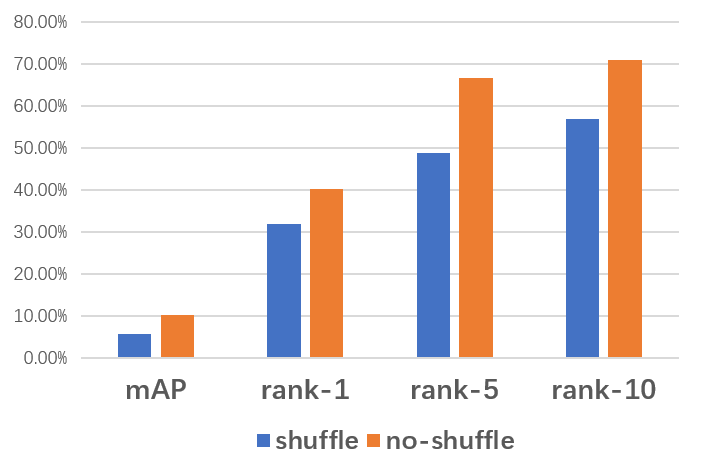} \\
	(a)  & (b)  & (c)  & (d)  \\
	\end{tabular}}
\caption{(a) The rank-n accuracy and mAP curve with different values of the manga frame threshold parameter $\sigma$ on the Manga109Face dataset.
(b) The rank-n accuracy and mAP curve with different values of the penalty factor parameter $\eta$ on the Manga109Face dataset. (c) The rank-n and mAP performance of the model on the face dataset when trained with the shuffle and no shuffle strategies. (d) The rank-n and mAP performance of the model on the body dataset when trained with the shuffle and no shuffle strategies.}\label{fig:parameters}
\end{figure*}

In addition, we also tested the comparison between the training strategy of shuffle and no shuffle on the datasets, and the test results for the face dataset and the body dataset are shown in Figure \ref{fig:parameters}(c) and Figure \ref{fig:parameters}(d). We observe that the training results using the no-shuffle strategy outperformed the shuffle strategy on both the face and body datasets. It indicates that the temporal order of manga character images is effective for clustering.

\subsection{Qualitative analysis}
We visualize the query results of some images from the face dataset in Figure \ref{fig:query} and displayed the rank-1 to rank-5 results of the search image (or query image).
Frame sequence indices below images express their temporal positions.
From this figure, we may further discover the detailed impact of our approach and discuss the following questions:

\textbf{Have spatial-temporal relationships helped clustering?}
As we can see, most negative results are far away from search images in the temporal dimension. From the perspective of method design, we shorten the distance of neighboring frame images while lengthening the distance of the images in the same frame. Therefore, we can avoid the clustering of remote images and images in the same frame as much as possible, so as to overcome the style limitation to a certain extent. The results of lines 1 and 2 in Figure \ref{fig:query} validate the effectiveness of our method.

\textbf{Has face-body combination helped clustering?}
Characters in lines 3 and 4 in the Figure \ref{fig:query} have similar clothes, and they are wrongly clustered together in the baseline without face-body combination. 
The face-body combination module considers two parts of the body, making it less possible to miss the signatures of characters. 
Hence, the experimental results corroborate our idea.


Moreover, we utilize t-SNE \cite{tsne} to visualize the feature embeddings of the clusters by plotting them to the 2D map, which is illustrated in Figure \ref{fig:cluster}. Although it is still difficult to distinguish some very similar images, our method has a satisfactory effect on the whole.

\section{Conclusion}
In this paper, we proposed a content-related unsupervised manga character re-identification method to achieve better clustering performance in the manga. 
The key is to utilize manga spatial-temporal relationships and combine face-body information of characters when clustered to tackle the problem of artistic expressions and style limitations in manga creation.
We designed a face-body graph to generate refined pseudo labels for each dataset.
Moreover, we modified the triplet loss with spatial-temporal relationships to the fine-tune network.
Experimental results on the Manga109 dataset validate the superiority of the proposed method.


\bibliography{tmm}
\bibliographystyle{plain}

\vfill

\end{document}